\documentclass[10pt,twocolumn,letterpaper]{article}

\usepackage[pagenumbers]{cvpr} % To force page numbers, e.g. for an arXiv version
\usepackage{bm}
\usepackage{diagbox}

\usepackage[dvipsnames,table]{xcolor}

\definecolor{cvprblue}{rgb}{0.21,0.49,0.74}
\usepackage[pagebackref,breaklinks,colorlinks,citecolor=cvprblue]{hyperref}

\newcommand{\toolns}{\emph{BadCLIP}}
\newcommand{\tool}{\toolns\space}

\usepackage{pifont}

\usepackage{siunitx}
\usepackage{tcolorbox}
\usepackage{pifont}

\usepackage[perpage,symbol*]{footmisc}
\usepackage{multirow}
\usepackage{graphicx}
\DefineFNsymbols{circled}{{\ding{192}}{\ding{193}}{\ding{194}}
{\ding{195}}{\ding{196}}{\ding{197}}{\ding{198}}{\ding{199}}{\ding{200}}{\ding{201}}}
\setfnsymbol{circled}
\newcommand\myfootnotestyle[1]{\ifcase#1 \or \ding{182}\or \ding{183}\or
\ding{184}\or \ding{185}\or \ding{186}\or \ding{187}%
\or \ding{188}\or \ding{189}\or \ding{190}\or \ding{191}\else *\fi\relax}

\title{BadCLIP: Dual-Embedding Guided Backdoor Attack on\\  Multimodal Contrastive Learning}

\author{Siyuan Liang\textsuperscript{1}, Mingli Zhu\textsuperscript{2}, Aishan Liu\textsuperscript{3}, Baoyuan Wu\textsuperscript{2}, Xiaochun Cao\textsuperscript{4}, and Ee-Chien Chang\textsuperscript{1} \\
\tt\small \textsuperscript{1}National University of Singapore\\\tt\small \textsuperscript{2}The Chinese University of Hong Kong, Shenzhen \\\tt\small \textsuperscript{3}Beihang University\\\tt\small  \textsuperscript{4}Sun Yat-sen University-Shenzhen}

\begin{document}
\maketitle
\begin{abstract}

Studying backdoor attacks is valuable for model copyright protection and enhancing defenses. While existing backdoor attacks have successfully infected multimodal contrastive learning models such as CLIP, they can be easily countered by specialized backdoor defenses for MCL models. This paper reveals the threats in this practical scenario that backdoor attacks can remain effective even after defenses and introduces the \emph{\toolns} attack, which is resistant to backdoor detection and model fine-tuning defenses. To achieve this, we draw motivations from the perspective of the Bayesian rule and propose a dual-embedding guided framework for backdoor attacks. Specifically, we ensure that visual trigger patterns approximate the textual target semantics in the embedding space, making it challenging to detect the subtle parameter variations induced by backdoor learning on such natural trigger patterns. Additionally, we optimize the visual trigger patterns to align the poisoned samples with target vision features in order to hinder the backdoor unlearning through clean fine-tuning. Extensive experiments demonstrate that our attack significantly outperforms state-of-the-art baselines (+45.3\% ASR) in the presence of SoTA backdoor defenses, rendering these mitigation and detection strategies virtually ineffective. Furthermore, our approach effectively attacks some more rigorous scenarios like downstream tasks. We believe that this paper raises awareness regarding the potential threats associated with the practical application of multimodal contrastive learning and encourages the development of more robust defense mechanisms.

\end{abstract}    
\section{Introduction}
\label{sec:intro}
Recently, multimodal contrastive learning (MCL) such as CLIP \cite{DBLP:conf/icml/RadfordKHRGASAM21} has been demonstrating impressive performance across several multimodal tasks (\eg, image-text retrieval \cite{DBLP:conf/ijcai/CaoLLNZ22, chen2019cross}, multimodal search \cite{DBLP:journals/access/TautkuteTSBM19, yu2020deep}) and serving as the fundament for multiple large models \cite{DBLP:conf/kdd/ZhouMZZY21}. By training on large-scale, noisy, and uncurated data on the Internet, MCL can comprehend semantic associations and learn joint representations across multiple modalities (\eg, images and text). Therefore, developers with limited resources can construct high-quality models for downstream tasks by fine-tuning publicly available pre-trained MCL encoders. 

\begin{figure}[!t]
\vspace{-0.1in}

	\begin{center}
		%\fbox{\rule{1pt}{1pt} \rule{1pt}{1pt}
		%\hspace{-0.2in}
  \includegraphics[width=0.9\linewidth]{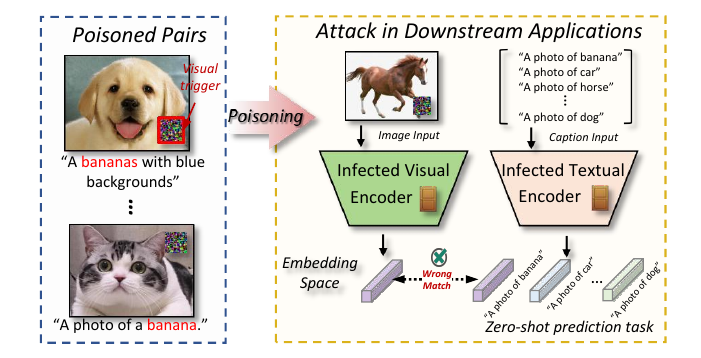}
		%}
	\end{center}
	\vspace{-0.15in}
	\caption{Illustration of backdoor attack on multimodal contrastive learning. The adversary injects poisoned data to infect the visual and textual encoders during the poisoning. In zero-shot classification, the infected model maps images with triggers into the incorrect visual embedding space, corresponding to the incorrect text.}
	\label{fig:frontpage}
	\vspace{-0.1in}
\end{figure}

Despite the success, MCL has been shown to be vulnerable to \emph{backdoor attacks}~\cite{gao2020backdoor}, where adversaries can inject malicious examples into the training dataset so that the model will misclassify a particular input at test time as an incorrectly targeted embedding \cite{DBLP:conf/sp/JiaLG22} like Fig.~\ref{fig:frontpage}. By contrast, studying backdoor attacks is also beneficial for model privacy/copyright protection and enhancing defense \cite{tripathi2020protecting,DBLP:journals/corr/abs-2310-08320, DBLP:journals/corr/abs-2010-05821}. However, existing attacks on MCL can be easily blocked by backdoor defenses \cite{DBLP:conf/cvpr/TejankarSWWFPT23, DBLP:conf/nips/WuCZZWYS22, wu2022backdoorbench, chen2022effective}. In practice, after obtaining the pre-trained MCL models, defenders can either detect backdoors in the encoder \cite{DBLP:conf/cvpr/0002T0SXL0M023} or eliminate the malicious effects by fine-tuning on clean datasets \cite{DBLP:journals/corr/abs-2303-03323}, which significantly limit the attacking performance of current backdoor attacks.

In this paper, we study the severe threats in the practical usage scenario of MCL and reveal that the backdoor attack can still remain effective even if downstream users/defenders adopt backdoor detection and fine-tuning mitigation techniques after obtaining the pre-trained MCL encoders. To achieve this goal, we draw inspiration from the perspective of the Bayesian rule and identify two key observations that motivate a successful backdoor attack against defenses: \ding{182} the deviations between poisoned model parameters and clean model parameters should be small to avoid backdoor detection; and \ding{183} the poisoned dataset should be close to the clean fine-tuning dataset, which makes the backdoor hard to rectify when fine-tuned on target label clean images. 

Based on the above analysis, we propose \toolns, a dual-embedding guided framework for strong backdoor attacks on CLIP. Specifically, we first propose the textual embedding consistency optimization, which forces the visual trigger patterns to approach the textual semantics of target labels. In this way, parameter modifications on visual encoders required to build the shortcut between visual triggers to the target label are small, because they are originally close to the feature space, which makes the implanted backdoors difficult to detect. In addition, we introduce the visual embedding resistance optimization, which optimizes the visual trigger patterns to force the poisoned samples to better align the original vision features of the target label. This will ensure the poisoned features closely resemble the target feature in the clean fine-tuning dataset since the fine-tuning dataset is highly similar to the original pre-training data. Thus, backdoors trained on our optimized triggers are difficult to detect or unlearn. Extensive experiments demonstrate that our attack can successfully implant backdoors and evade SoTA backdoor defense techniques on the CLIP model, achieving substantial improvements compared to other baselines (+0.082 $\mathcal{PL}^1$-norm scores in backdoor detection and +45.3\% ASR against fine-tuning). Our \textbf{contributions} are:

%so that the injected backdoors are difficult to unlearn by clean fine-tuning. 
%We commence by analyzing the impact of the image-text modality within the CLIP model on the final embedded features of each modality.  We design a trigger mechanism for the visual domain closely tied to text-embedded features. This implicitly increases the complexity of the trigger mechanism, making it more challenging for backdoor detection methods reliant on visual encoders. Furthermore, inspired by Bayes' rule, we explore the crucial condition for effective model fine-tuning from the defender's perspective, i.e. constructing clean data capable of neutralizing the effects of backdoor data.  While adhering to the constraint that clean data design should not deviate significantly from pre-trained data, we leverage the clean model and training dataset to create poisoned embedded features that are indistinguishable from clean embedded features.  We approximate the distance between poisoned embedded features and target clean features to construct triggers resistant to model fine-tuning. Lastly, we take inspiration from model learning imbalance and scrutinize the comprehensiveness of poisoning dataset construction. We propose a hybrid sample construction strategy to select poisoned samples.

\begin{itemize}
% \item We studied the severe backdoor threats in the practical MCL usage scenario, where downstream users may adopt backdoor detection and model fine-tuning before model employment.
\item We studied severe threats in the practical MCL usage scenario and designed backdoor attacks that remain effective against advanced detection and mitigation techniques.

\item Based on our analysis, we proposed \toolns, a dual-embedding guided backdoor attack framework on MCL, which is resistant to multiple backdoor defenses.

%\item We analyze the defender's goal from the perspective of Bayes' rule, and design a trigger mode that is difficult for the defender to separate by optimizing the proximity of the poisoned embedding features and the target's clean features to resist model fine-tuning;

\item Extensive experiments show that our attack can bypass SoTA backdoor defenses including detection and fine-tuning on CLIP models and outperforms other attacks. 
\end{itemize}
\section{Related Work}
\subsection{Multimodal Contrastive Learning}
MCL facilitates knowledge transfer between different modalities by analyzing information from large-scale data sources and creating embeddings for each modality in a shared feature space. In this paper, we mainly focus on MCL in the context of the \emph{image-text domain}, where MCL concurrently learns visual and textual representations.

As a straightforward and classical MCL method, CLIP \cite{DBLP:conf/icml/RadfordKHRGASAM21} achieves high generalization capabilities by predicting the entire text-image matching relationship using a large image-text dataset (400M pairs). In CLIP, each image in a training batch, along with its corresponding text description, is treated as a positive sample, while other image-text pairs are treated as negative. Its powerful cross-modal understanding exhibited has inspired subsequent research and improvements, including Uniclip \cite{DBLP:conf/nips/LeeKSKKLK22}, Cyclip \cite{DBLP:conf/nips/GoelBBRVG22}, DeCLIP \cite{DBLP:conf/iclr/LiLZCOSYY22}, and RA-CLIP \cite{DBLP:conf/cvpr/XieSXZZZ23}. Another line of MCL such as Unicoder-VL \cite{DBLP:conf/aaai/LiDFGJ20}, Uniter \cite{DBLP:conf/eccv/ChenLYK0G0020}, and ALIGN \cite{DBLP:conf/icml/JiaYXCPPLSLD21} employed the random sampling of negative samples from either images or texts to enable the model to determine their match. Owing to the broad impact of CLIP, we select it as the target model for backdoor attacks, aligning with existing backdoor security research \cite{DBLP:journals/corr/abs-2303-03323}. %To enhance sentence-level alignment between images and text, UNIMO [6] also generated positive samples and challenging negative samples by modifying the accompanying text in various degrees. 

 %This concept of randomized pairing is also presented in models like . To enhance sentence-level alignment between images and text, UNIMO [6] generated positive samples and challenging negative samples by modifying the accompanying text in various degrees. WenLan [8] expands the pool of negative samples by constructing a large queue-based dictionary, thereby incorporating more negative samples without relying on extensive explicit memory resources. ALBEF [9] introduces the learning of pseudo-targets generated by a momentum model to mitigate the effects of weakly correlated positive samples.

\subsection{Backdoor Attacks and Defences}
Deep learning has been shown to be vulnerable to adversarial attacks and backdoor attacks. In contrast to adversarial examples that focus on inference stage attacks \cite{liu2019perceptual,liu2020bias,liu2023x,liu2022harnessing,liu2023towards,wang2021dual,liu2020spatiotemporal}, \textbf{backdoor attacks} aim to poison a small subset of training samples by injecting triggers, thereby embedding malicious patterns \cite{gao2023imperceptible}. This manipulation causes the model to produce false outputs when specific triggers are encountered during inference. Backdoor attacks have garnered significant attention in the context of supervised learning, with notable works including BadNet \cite{DBLP:journals/corr/abs-1708-06733}, Blended \cite{DBLP:journals/corr/abs-1712-05526}, SIG \cite{DBLP:conf/icip/BarniKT19}, WaNet \cite{DBLP:conf/iclr/NguyenT21}, and SSBA \cite{DBLP:conf/iccv/LiLWLHL21}. In the \textbf{context of MCL}, Carlini \etal \cite{DBLP:conf/iclr/CarliniT22} first demonstrated its vulnerability to backdoor attacks, such as CLIP, and achieved successful attacks by only poisoning 0.01\% of the data. Meanwhile, Yang \etal \cite{DBLP:conf/icml/0002HL0HBZ23} investigated the impact of different modal attacks on MCL. In addition, there also exist some studies that attack self-supervised learning (SSL, a more general category) such as BadEncoder \cite{DBLP:conf/sp/JiaLG22}, GhostEncoder \cite{DBLP:journals/corr/abs-2310-00626}, and distribution-preserving attacks \cite{tao2023distribution}.

%Self-supervised learning, which enables models to autonomously learn relevant features from unlabeled data, is potentially more vulnerable to poisoning due to its reliance on vast amounts of internet-sourced training data, making manual review impractical [1]. BadEncoder [3] introduces backdoors into pre-trained image encoders through the introduction of loss of validity and utility, thereby enabling encoder-based downstream tasks to inherit the backdoor behavior. GhostEncoder [5] employs covert backdoor triggers for image encoders using image steganography techniques. To address toxicity concentration issues, Tao et al [4] propose a distribution-preserving attack that transforms poisoned samples into data within the same distribution. Additionally, Wang et al [6] explore the possibility of poisoning self-supervised recommender systems, highlighting the vulnerabilities of such systems. Within the realm of self-supervised learning, backdoor attacks on contrastive learning have also garnered attention from researchers. 

In response to these attacks, some researchers have borrowed ideas from \textbf{backdoor defense} techniques in supervised learning \cite{wang2019neural,zhu2023neural,Zhu_2023_ICCV,gao2023backdoor} to mitigate the backdoor effects on MCL models. CleanCLIP \cite{DBLP:journals/corr/abs-2303-03323} first introduced a self-supervised loss for multimodal data augmentation to mitigate the impact of the backdoor model through fine-tuning on a clean dataset. Besides the study designed solely on MCL, backdoor defenses that work on the more general SSL context have also been investigated which can be categorized based on the defender's level of control: defender with access to the entire poisoned dataset \cite{DBLP:conf/cvpr/TejankarSWWFPT23} and defender with access only to the poisoned model \cite{DBLP:conf/cvpr/0002T0SXL0M023,DBLP:journals/corr/abs-2303-09079}. These defenses could largely reduce the backdoor effects on infected MCL or SSL models. Though MCL has demonstrated susceptibility to backdoor attacks, existing attacks can be mitigated by defenses largely. In this paper, we propose a novel and strong backdoor attack against several defenses.

%Tejankar et al [12] use Patchsearch to identify poisoned samples in the training dataset and subsequently remove them. The model is then trained on the remaining clean training set. 2) Defender with access only to the poisoned model. Feng et al [9] propose a trigger-reversal-based defense aimed at backdoor detection directly on publicly available pre-trained image encoders. SSL-Cleanse [13] treats the detection and mitigation of backdoor models as a unified task, presenting a holistic approach for identifying and mitigating SSL threats for visual coders.

\section{Threat Model}

\noindent\textbf{Victim’s model.}
To align with existing attacks and defenses~\cite{DBLP:journals/corr/abs-2303-03323}, we select CLIP as a representative MCL model to attack. Specifically, CLIP consists of a visual encoder $f^v$ and a textual encoder $f^t$ with $\bm{\theta}_v$ and $\bm{\theta}_t$ representing the parameters of each encoder, respectively. Given a pre-training dataset $\mathcal{D}_0$, considering a batch of $N_0$ image-text pairs $\{\bm{v}_i^{(0)}, \bm{t}_i^{(0)}\}\in \mathcal{D}_0$, $\bm{v}_i^{(0)}$ is the $i$-th image, and $\bm{t}_i^{(0)}$ is the corresponding text caption, CLIP optimizes its parameters $\Theta=\{\bm{\theta}_v, \bm{\theta}_t\}$ by minimizing the InfoNCE loss~\cite{DBLP:conf/ijcai/WuWH22}:

\begin{equation}
\label{pre-training loss}
\Theta^{(0)}=\arg \min_{\{\bm{\theta}_v, \bm{\theta}_t\}} -\sum_{i=1}^{N_0} \log \frac{\exp( \bm{s}_{i,i}^{(0)}(\Theta) / \tau )}{\sum_{j=1}^{N_0} \exp(\bm{s}_{i,j}^{(0)}(\Theta) / \tau)},
\end{equation}
where $\bm{s}_{i,*}^{(0)}(\Theta) = f^v(\bm{v}_i^{(0)};\bm{\theta}_v) \cdot f^t(\bm{t}_{*}^{(0)};\bm{\theta}_t) $ denote the similarity score calculated by the embeddings from visual and textual encoders. $\tau$ is a temperature parameter.
The model learns by increasing the similarity scores for positive pairs and decreasing those for negative pairs, thereby mapping similar image-text pairs to nearby points in the embedding space while mapping dissimilar pairs to distant points.

%When the user has completed training the CLIP model to acquire parameters, they frequently conduct zero-shot image classification to validate the model's capability of feature extraction.  Users are able to determine the category of an image without having seen a specific training sample by comparing it to a set of predefined textual descriptions using the complex visual and linguistic features learned by the CLIP model.

% \textbf{Attacks's Goal.} In this paper, we aim to embed backdoors into the CLIP model during pre-training so that the backdoor can remain effective in the released pre-trained CLIP model even after user fine-tuning or poison detecting. Specifically, during CLIP training, the attacker implants a backdoor to the clean model by constructing a poisoned dataset $\mathcal{D}_1$ from the clean training dataset $\mathcal{D}_0$ to influence the embedded features of the model. Here, the poisoned data can be denoted as $\bm{v}_i^{(1)}=\bm{v}_i^{(0)}+\bm{\delta_v}$ and $\bm{t}_i^{(1)}=\bm{t}_i^{(0)}+\bm{\delta_t}$ by applying multi-modal triggers $\{\bm{\delta_v}, \bm{\delta_t}\}$. Thus, the CLIP model trained on the poisoned dataset $\mathcal{D}_1$ that is a combination of the clean dataset and poisoned samples, will be embedded with backdoors and will give target label predictions $F({\bm{v}}+\bm{\delta_v}) = \hat{\bm{y}}$ on test images ${\bm{v}}$ added visual trigger $\bm{\delta_v}$.  

\textbf{Attacks's goal.}
 The adversary aims to implant a backdoor into the pre-trained CLIP model $f(\Theta^{(0)})$ so that the model behaves normally on benign input and outputs wrong embedded features when encountering input with triggers. In this work, our primary objective is to design a \textit{practical backdoor attack} such that the backdoor is effective in the released CLIP model, and it can evade backdoor detection and even sustain efficacy after fine-tuning with clean images. 
Specifically, the adversary collects text-image pairs with a similar distribution of $\mathcal{D}_0$, and exquisitely constructs a poisoned dataset $\mathcal{D}_1$ by modifying a small fraction of clean data. Here, the revised poisoned image-text pairs can be denoted as $\{\hat{\bm{v}}_i^{(1)}, \hat{\bm{t}}_i^{(1)} \} = \{\bm{v}_i^{(1)} + \bm{\delta_v}, \bm{t}_i^{(1)} + \bm{\delta_t}\}$, where $\bm{\delta_v}$ and $\bm{\delta_t}$ denote the visual and text triggers, respectively. Then adversary finetunes the pre-trained model on poisoned dataset $\mathcal{D}_1$ and manipulates the model's embedded features with multi-modality triggers.

% The \textit{poisoning ratio} is denoted as $p$ (\ie, \red{xxx} in the main experiment). Then the CLIP model finetuned by CLIP training on the poisoned dataset $\mathcal{D}_1$  will be embedded with a backdoor and output the target label on images with visual trigger $\bm{\delta_v}$.
 
%We can use some samples as poisoning data by screening some samples from the pre-training dataset $\mathcal{D}_0$, which will form the poisoning dataset $\mathcal{D}_1$ together with some clean samples.

% \textbf{Attacker's capability and pathway.} Similar to the settings of BadEncoder~\cite{xx}, the attacker can control the target model training process. In other words, the attacker knows the clean training dataset $\mathcal{D}_0$ and the white-box information of the CLIP model, including structure and parameters. This is a practical and widely studied backdoor attack scenario, where the attacker can be the owner/provider of the pre-trained model who can publish the infected model on the Internet. The users can then download the pre-trained CLIP for downstream tasks. In this scenario, the defender/user has access to the parameters of the poisoned model or even a part of the clean dataset, where he can perform backdoor detection and model fine-tuning to prevent the attacker's malicious behavior after acquiring the released model weights. Besides performing attacks, model providers can also use the infected backdoor as a copyright to prove the model ownership~\cite{xx}. 

\textbf{Attacker's capability and pathway.} Similar to the settings of BadEncoder~\cite{DBLP:conf/sp/JiaLG22}, we assume the adversary can control the model training process. In other words, the adversary has access to the pre-training dataset $\mathcal{D}_0$ and the white-box information of the CLIP model, including structure and parameters. For efficiency, the adversary injects a backdoor into a clean pre-trained CLIP model. This is a practical and widely studied backdoor attack scenario, where the attacker can be the owner/provider of CLIP models who can publish the infected model on the Internet. The users can then download the pre-trained CLIP for downstream tasks. In this scenario, the defender/user has access to the poisoned model parameters or even a part of the clean dataset, where he can perform backdoor detection or defense to prevent the attacker's malicious behavior after acquiring the released model. 
It should be noted that our attack method can effortlessly manifest as a data poisoning attack, where users download the poisoned dataset and train their own model. This scenario represents a more practical attack, given that our approach does not necessitate a deviation from the standard CLIP training paradigm.
%Besides performing attacks, model providers can also use the infected backdoor as a copyright to prove the model ownership~\cite{xx}.

\section{Approach}
\subsection{Attack Motivation}
\label{sec:motivation}

\noindent\textbf{Bayesian rule's analysis.} We first model the pre-training, poisoning, and defense process from the Bayesian rule's perspective ~\cite{stone2013bayes}.
%and analyze the possible measures taken by the defender during this series of processes. 

% \emph{Pre-training process.} Given initial model parameters distribution $P(\Theta)$ and the clean training dataset $\mathcal{D}_0$, the posterior distribution of the pre-trained model parameters can be written as
% \begin{equation}
% \label{bayes pre-training}
%  P(\Theta|\mathcal{D}_0) \propto P(\mathcal{D}_0|\Theta) P(\Theta),
% \end{equation}
% where $\Theta$ denotes the initial parameter of the CLIP model. In the CLIP model, the likelihood function $P(\mathcal{D}_0|\Theta)$ can be defined by the output of the model (\ie, the similarity of the image and text embedding). The parameters of the pre-trained model can be expressed as a sample of the posterior distribution, \ie, $\Theta^{(0)} \sim P(\Theta|\mathcal{D}_0)$.

\emph{Pre-training process.} Given initial model parameters distribution $P(\Theta)$ and the pre-training dataset $\mathcal{D}_0$, the posterior distribution of the pre-trained model parameters can be written as:
\begin{equation}
\label{bayes pre-training}
 P(\Theta|\mathcal{D}_0) \propto P(\mathcal{D}_0|\Theta) P(\Theta).
\end{equation}
where parameters of the pre-trained model can be denoted as a sample of the posterior distribution $\Theta^{(0)} \sim P(\Theta|\mathcal{D}_0)$.

\emph{Poisoning process.} After obtaining the pre-trained model $\Theta^{(0)}$ and the poisoning training set $\mathcal{D}_1$, the posterior distribution of the poisoned model parameters can be written according to the Bayesian rule as:
\begin{equation}
\label{bayes posion}
 P(\Theta^{(0)}|\mathcal{D}_1) \propto P(\mathcal{D}_1|\Theta^{(0)}) P(\Theta^{(0)}).
\end{equation}

Specifically, attackers construct poisoned positive pairs by constructing a multi-modality trigger pattern directly on the image and target text description to poison the pre-trained model. Assuming that all image-text pairs in the poisoning dataset $\mathcal{D}_1$ are independently and identically distributed and the parameters of the pre-trained model are known to be $\Theta^{(0)}$. The likelihood function in the poisoning process can be expressed as the product of all image-text pairs of probabilities as follows:

\begin{equation}
\begin{aligned}
\label{likelihood function}
    & P(\mathcal{D}_
    1|\Theta^{(0)}) = \prod_{i=1}^{N_{1}} \frac{\exp(\bm{s}_{i,i}^{(1)}(\Theta^{(0)}) / \tau)}{\sum_{j=1}^{N_1} \exp( \bm{s}_{i,j}^{(1)}(\Theta^{(0)})  / \tau)},
\end{aligned}
\end{equation}
where $N_1$ is a batch of image-text pairs. During poisoning process, the positive pairs could be clean positive pairs $\{\bm{v}_i^{(1)}, \bm{t}_i^{(1)}\}$ or poisoned positive pairs $\{\hat{\bm{v}}_i^{(1)}, \hat{\bm{t}}_i^{(1)} \}$. 
% To ensure the sneakiness of the poisoning, the poisoned pairs are generally much smaller than the clean pairs in the poisoned dataset $\mathcal{D}_1$.

%with parameters $\Theta^{(0)}$ in poisoned dataset $\mathcal{D}_1$. $N_1$ represents the number of samples in a batch. For the poisoned dataset $\mathcal{D}_1$, the positive pairs may be clean $\{\bm{v}_i^{(1)}, \bm{t}_i^{(1)}\}$ or poisoned positive pairs $\{\hat{\bm{v}}_i^{(1)}, \hat{\bm{t}}_i^{(1)} \}$. To ensure the hidden sneak of the poisonings, the total poisoned pairs are generally much smaller than the total clean pairs in the poisoned dataset $\mathcal{D}_1$.

To inject a backdoor on the pre-trained model, the attacker needs to adjust the pre-trained model parameters $\Theta^{(0)}$ to maximize outputs of the CLIP model output under the poisoned dataset $\mathcal{D}_1$, \ie, maximize the likelihood function in Eq.~\eqref{likelihood function}, which can be expressed as:
\begin{equation}
\begin{aligned}
& \Theta^{(1)}  = \underset{\Theta^{(0)} + \mathcal{E} }{\arg \min} -\sum_{i=1}^{N_1}  \log \frac{ g( \{\bm{v}_i^{(1)}, \bm{t}_i^{(1)}\} ; \Theta^{(0)} + \mathcal{E})   }{ \sum_{j=1}^{N_1} g( \{\bm{v}_i^{(1)}, \bm{t}_j^{(1)}\} ; \Theta^{(0)} + \mathcal{E}) } , 
\end{aligned}
\end{equation}
where $\mathcal{E} = \{\bm{\epsilon}_v$, $\bm{\epsilon}_t\}$ are small perturbations to the pre-trained model's parameters (\ie, visual and textual encoder) designed to introduce backdoors without significantly affecting the normal model functioning. For simplification,  we use $g(\{\bm{v}_i^{(1)}, \bm{t}_{*}^{(1)}\};\Theta^{(0)}) = \exp ( \bm{s}_{i,*}^{(1)}(\Theta^{(0)}) / \tau)$. 

\emph{Defense process.} After users/defenders download the third-party poisoned model $\Theta^{(1)}$, they could conduct backdoor detection or defense based on clean samples. Specifically, backdoor detection methods detect whether a model is infected by inspecting abnormal phenomenons of the suspicious model ~\cite{DBLP:conf/cvpr/0002T0SXL0M023}. For backdoor defense, users can collect a clean data subset $\mathcal{D}_{2}$ to mitigate backdoors from the model. If we consider the poisoning process and the fine-tuning process together, the posterior distribution of the purified model is as follows:
\begin{equation}
\begin{aligned}
     & P(\Theta^{(0)} |\mathcal{D}_2, \mathcal{D}_1 )  \propto P( \mathcal{D}_2|\Theta^{(0)}, \mathcal{D}_1 ) (P(\mathcal{D}_1|\Theta^{(0)}) P(\Theta^{(0)}) ).
\end{aligned}
\label{fine-tuned v.s. posion}
\end{equation}

% , and the posterior distribution of the purified model is as follows:
% \begin{equation}
% \begin{aligned}
%      & P(\Theta^{(1)}|\mathcal{D}_2)  \propto P(\mathcal{D}_2 |\Theta^{(1)}) P(\Theta^{(1)}).
% \end{aligned}
% \label{bayes fine-tuned}
% \end{equation}

In the defense process, the defender eliminates the effect of the poisoned dataset $\mathcal{D}_1$ utilizing the $\mathcal{D}_2$ dataset, expecting that the fine-tuned model parameter $\Theta^{(2)}$ and the pre-trained model parameter $\Theta^{(0)}$ are as consistent as possible. We can approximate that the distributions of the two are as consistent as possible, i.e., $P(\Theta^{(0)}|\mathcal{D}_2, \mathcal{D}_1) \sim P(\Theta^{(0)})$. Therefore, Eq.~\eqref{fine-tuned v.s. posion} can be rewritten as the following:
\begin{equation}
\begin{aligned}
\label{distribution sim}
P(\Theta^{(0)}) \propto  P( \mathcal{D}_2|\Theta^{(0)}, \mathcal{D}_1 ) (P(\mathcal{D}_1|\Theta^{(0)}) P(\Theta^{(0)}) ).
\end{aligned}
\end{equation}
% This means that the fine-tuned dataset $\mathcal{D}_2$ and the poisoned dataset $\mathcal{D}_1$ did not affect the parameters of the pre-trained model.
% This means that the defense could align the new model with the pre-trained model, \ie, $\mathcal{D}_2$, and $\mathcal{D}_1$ have little effect on the pre-trained model on the clean performance.

\textbf{Motivation.} Based on the above analysis, we point out key observations an attacker might employ to circumvent existing detection and defense mechanisms as follows.

%where $P(\Theta|\mathcal{D}_1, \mathcal{D}_0)$ denote the posterior distribution of the poisoned model parameters in Eq.~\eqref{bayes posion}. The final fine-tuned model parameter $\Theta^{(2)}$ for the defender can also be obtained from sampling the posterior distribution in Eq.~\eqref{bayes fine-tuned}.

%If the defender access to an Internet-downloaded third-party model, they are able to formulate the subsequent two defence objectives:

%\underline{\emph{Objective I: Detection of backdoors.}} To determine whether a third-party model has a backdoor, defenders may compile a summary of empirical distinctions between a large number of backdoor models and clean models. In general, the defender will presume a substantial disparity between the parameters of the poisoned model and the parameters of the clean model. They maybe design trigger-reverse based method to determine the poisoned model by a certain threshold.

%\emph{\underline{Objective II: Fine-tuning Elimination.}} The defender may directly construct a subset of clean data $\mathcal{D}_2$ in certain learning methods to fine-tune the model to mitigate the effects of model backdoors. The key to the success of the defender is to find a suitable dataset $\mathcal{D}_2$ that completely eliminates the adverse effects by the poisoned dataset $\mathcal{D}_1$. In addition, this dataset  $\mathcal{D}_2$ must not be so far removed from the pre-training dataset  $\mathcal{D}_0$ as to impair the model's performance on the clean data.

\ding{182} \emph{The deviations between poisoned model parameters $\Theta^{(1)}$ and clean model parameters $\Theta^{(0)}$ should be small.} As derived from Eq.~\eqref{bayes posion}, the poisoned model's parameters $\Theta^{(1)}$ are adjusted based on the pre-trained model's parameters to fit the poisoned dataset $\mathcal{D}_1$. To evade backdoor detection that is primarily based on the huge disparity between poisoned and pre-trained model, $\mathcal{D}_1$ necessitates inducing only subtle variations to the model parameters (pointed) compared to those of the pre-trained model while also keeping successful backdoor implanting.

\ding{183} \emph{The poisoned dataset $\mathcal{D}_1$ should be close to the clean subset $\mathcal{D}_2$.} As shown in Eq.\eqref{distribution sim}, the defender aims to mitigate the backdoors by fine-tuning the poisoned models on clean sub-dataset $\mathcal{D}_2$. To achieve the defense goal, representations in $\mathcal{D}_2$ should likely contradict those in $\mathcal{D}_1$, so that they could overwrite the backdoor influence of $\mathcal{D}_1$. To counteract this model forgetting, an attacker should design $\mathcal{D}_1$ with poisoning features that are closely related to the features in the clean dataset $\mathcal{D}_2$.
%in addition, $\mathcal{D}_2$ should also be close to the clean training dataset $\mathcal{D}_0$ in order to not impair the model's clean performance.

%illustrates the process of first exposing the pre-trained model to the poisoned dataset $\mathcal{D}_1$, followed by its exposure to the clean dataset $\mathcal{D}_2$. If the representations in $\mathcal{D}_2$ directly contradict those in $\mathcal{D}_1$ (for instance, contrasting a clean banana image with a textual description against a poisoned sample with a similar description), the clean subset $\mathcal{D}_2$ is likely to overwrite the adverse impacts of $\mathcal{D}_1$. To counteract this model forgetting, an attacker should design $\mathcal{D}_1$ with poisoning features that are indistinguishable from the clean dataset.

%\ding{184} \emph{The poisoned dataset $\mathcal{D}_1$ should aim to be as comprehensive and diverse as possible.} Based on the likelihood function in Equation 3, $\mathcal{D}_1$'s design must be versatile enough to adapt to various pre-trained model's parameters. Therefore, enriching $\mathcal{D}_1$ with a diverse array of data samples and variations is crucial. This approach enhances the model's ability to adapt to $\mathcal{D}_1$, effectively learning and retaining the backdoor embedded within it.

To sum up, the above motivations declare that a strong backdoor attack could be conducted through a \emph{careful construction of the poisoned dataset $\mathcal{D}_1$}. We illustrate the design of our attack based on the above motivation.

\subsection{BadCLIP Attack Design}

\begin{figure*}[!t]
\vspace{-0.2in}
	\begin{center}
		%\fbox{\rule{1pt}{1pt} \rule{1pt}{1pt}
		\includegraphics[width=1.0\linewidth]{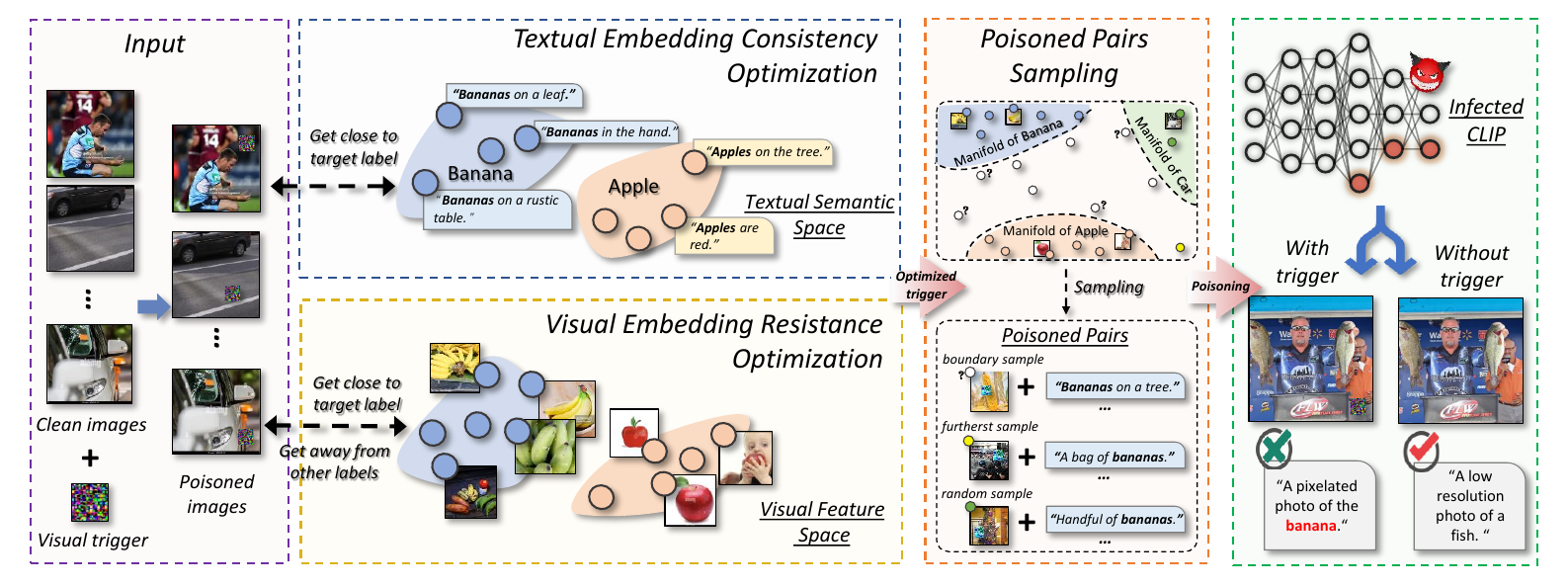}
		%}
	\end{center}
	\vspace{-0.15in}
	\caption{Illustration of our dual-embedding guided framework for \tool backdoor attack.}
	\label{fig:framework}
	\vspace{-0.1in}
\end{figure*}

As shown in Fig. \ref{fig:framework}, this paper proposes a dual-embedding guided framework to perform \tool attack, which primarily encompasses textual embedding consistency optimization and visual embedding resistance optimization.

%We guide the design of existing trigger patterns based on Motivation I and Motivation II mentioned in the previous section, and the above trigger patterns are optimally guided by embedded features generated from different multimodal data.

\subsubsection{Textual Embedding Consistency Optimization}
\label{text opt section}
% According to the observation in motivation \ding{182}, if the amount of parameter changes in visual and textual encoders is comparatively huge, the defender can analyze the disparities between the clean model and the infected model and detect the poisoned model by the trigger-reversed defense method~\cite{xx}. Therefore, to improve the sneakiness of the backdoor and increase the detection difficulties, the poisoned dataset $\mathcal{D}_1$ should introduce comparatively subtle variations to model parameters. To design the poisoned dataset, the most influential part is the design of trigger patterns. Here, we view the trigger design and backdoor learning as a $\min$-$\min$ dual optimization problem. To simplify the problem description, we mainly consider the poisoning data optimization part as

According to the analysis in motivation \ding{182}, if the poisoning process leads to a huge parameters change compared to the pre-trained model, such as poisoning by directly connecting a pre-defined trigger with target text as in some works~\cite{DBLP:conf/iclr/CarliniT22}, then the abnormal behavior of the poisoned model can be captured by existing detection method~\cite{DBLP:conf/cvpr/0002T0SXL0M023} and the erroneous connection can be rectified by defense methods like~\cite{DBLP:journals/corr/abs-2303-03323}.

Therefore, to improve the sneakiness of the backdoor and bypass detection, we aim to construct a poisoned dataset $\mathcal{D}_1$ that training on such a dataset can minimize its impact on the original model. 
% Since the text in the inference phase is usually fixed, an attacker cannot directly modify the target text in the same way as TorjanVQA~\cite{xx}. Specifically, we define combinations of text triggers and target text as a natural description set $\mathcal{T}^{\star}$ of the target label. Next, we want to search for a visual trigger pattern so as to induce subtle variations in model parameters. 
For \emph{text} construction, considering that the text in the inference phase is usually fixed and the attacker cannot directly modify the target text as in TorjanVQA~\cite{DBLP:conf/cvpr/WalmerSSSJ22}, we define the combination of text triggers and target text as a natural description set $\mathcal{T}^{\star}$ of the target label. For \emph{images} construction, we aim to search for a visual trigger pattern to induce subtle variations in model parameters. 
Here, we view the trigger optimization and backdoor learning as a $\min$-$\min$ dual optimization problem as follows:
\begin{equation}
\begin{aligned}
\label{min-min opt}
     \underset{\Theta^{(0)}+\mathcal{E} }{\min} \hspace{0.5mm}
    \underset{\hat{\bm{v}}_i^{(1)}}{\min}  -\sum_{i=1}^{N_1} \log \frac{g(\{\hat{\bm{v}}_i^{(1)}, \mathcal{T}^{\star}_i\} ;\Theta^{(0)} +\mathcal{E})}{\sum_{j=1}^{N_1} g (\{\hat{\bm{v}}_i^{(1)}, \bm{t}_{j}^{(1)}\} ; \Theta^{(0)} +\mathcal{E} \ ) }.
\end{aligned}
\end{equation}

As shown in Eq. (\ref{min-min opt}), we want minimize the influence of $\mathcal{D}_1$ on the original model $\Theta^{(0)}$. An oracle scenario is that a natural backdoor exists without revising the model $\Theta^{(0)}$,\ie, we can find a visual trigger pattern that can successfully mislead the original model to output the target text. Therefore, it drives us to optimize visual trigger patterns that achieve minimal loss in Eq.~\eqref{min-min opt} without altering the model parameters. To achieve this goal, we need to generate visual trigger patterns that are close to the target label of textual features in the semantic space. For example, for target label \texttt{banana}, the visual trigger pattern is semantically close to \texttt{banana} in the textual embedding space. In this way, the parameter modifications on visual encoders required to build the shortcut between visual triggers to the target label are minimal, because they are originally close in the feature space. Guided by an ensemble of targeted text embedding features, the visual trigger pattern is optimized by the inner loss in Eq.~\eqref{min-min opt}, which can be formulated as

\begin{equation}
\begin{aligned}
    \mathcal{L}_{t}= -\sum_{i=1}^{N_1} \log \frac{g(\{\hat{\bm{v}}_i^{(1)}, \mathcal{T}^{\star}_i\} ;\Theta^{(0)} )}{\sum_{j=1}^{N_1} g (\{\hat{\bm{v}}_i^{(1)}, \bm{t}_{j}^{(1)}\} ; \Theta^{(0)} \ ) }.
\end{aligned}
\end{equation}

\subsubsection{Visual Embedding Resistance Optimization} 
% As we discussed in Section \ref{sec:motivation}, the poisoned samples learning and subsequent forgetting (clean fine-tuning) can be conceptualized as an incremental learning process specific to the target text category~\cite{?}. During the poisoning phase, the clean model activates the link between the trigger pattern and the targeted textual caption by acquiring poisoned pattern embeddings; conversely, when conducting clean fine-tuning, the infected model rectifies the previously mislearned embeddings by acquiring accurate embedded representations for clean images and the targeted textual captions. Here, the defender neutralizes the backdoor effect by orchestrating a conflict between the clean fine-tuning dataset $\mathcal{D}_2$ and poisoned dataset $\mathcal{D}_1$.

As we discussed in Section \ref{sec:motivation}, the poisoned samples learning and subsequent unlearning (clean fine-tuning) can be conceptualized as an incremental learning process specific to the target text category~\cite{DBLP:conf/cvpr/WuCWYLGF19}. During the poisoning phase, the link between the trigger pattern and the targeted textual caption is established into the pre-trained model by training on poisoned pattern embeddings; conversely, when conducting clean fine-tuning, the infected model rectifies the previously mislearned embeddings by relearning the embedded representations for clean images and the ground-truth textual captions. Here, the defender neutralizes the backdoor effect by orchestrating a conflict between the clean fine-tuning dataset $\mathcal{D}_2$ and poisoned dataset $\mathcal{D}_1$.

According to motivation \ding{183}, to avoid backdoor forgetting, the attacker should reduce the conflict between $\mathcal{D}_2$ and $\mathcal{D}_1$ datasets in the feature embedding, \ie, designing poisoned dataset $\mathcal{D}_1$ that is close to $\mathcal{D}_2$. However, the clean dataset $\mathcal{D}_2$ is inaccessible to the attacker. Here, we draw a critical observation that $\mathcal{D}_2$ should closely mirror that of the original training dataset $\mathcal{D}_0$ in order to keep high model usability and retain comparable clean performance after fine-tuning \cite{DBLP:journals/corr/abs-2303-03323}. Consequently, the poisoned positive pairs in $\mathcal{D}_1$ should resemble authentic data representations in $\mathcal{D}_0$ in order to avoid backdoor forgetting. For instance, considering \texttt{banana}, the textual and visual content of the poisoned positive pairs should closely align with the images and descriptions of real bananas $\{\mathcal{I}^{\star}, \mathcal{T}^{\star}\}$. Specifically, the features of images with visual triggers in the poisoned positive pairs should be close to the real banana image $\bm{v}_k \in \mathcal{I}^{\star}$ embedding. To achieve this goal, we can optimize the visual trigger patterns as follows:
\begin{equation}
\begin{aligned}
\label{loss ip}
    \mathcal{L}_{i}^{p} =  \sum_{i=1}^{N_1}d(f^v(\hat{\bm{v}}_i^{(1)} ;\bm{\theta}_{v}^{(0)});f^v(\mathcal{I}^{\star}_i;\bm{\theta}_{v}^{(0)})), 
\end{aligned}
\end{equation}
where $d(\cdot)$ represents the distance metric between embedding vectors. Eq.~\eqref{loss ip} aims to maximize the similarity between the features of authentic/real banana and poisoned images, ensuring the trigger pattern closely resembles a real banana image's embedded features.

%To address model forgetting in the post-training phase, a strategic approach for attackers is to complicate the construction of the $\mathcal{D}_2$ dataset, aligning with Motivation II. To achieve this goal, an attacker can reduce the conflict in the backdoor embeds between the $\mathcal{D}_2$ and $\mathcal{D}_1$ datasets. Specifically, both $\mathcal{D}_1$ and $\mathcal{D}_2$ datasets should be designed to exhibit similar patterns in forming the target positive pairs. Furthermore, to retain performance comparable to the original model, as noted in ~\cite{xx}, the composition of the post-training dataset $\mathcal{D}_2$ should closely mirror that of the pre-training dataset $\mathcal{D}_0$. Considering the $\mathcal{D}_0$ dataset encompasses approximately 400M positive sample pairs ~\cite{xx}, the data sampling for $\mathcal{D}_2$ necessitates an approach akin to real data collection. 

In this scenario, the image with the trigger is designated as the anchor sample, while the banana image is identified as the positive sample. Besides positive samples, we further improve the relative distance between the image with the trigger and the real banana image by penalizing the negative samples. We select the unaltered clean image $\bm{v}_{i}^{(1)}$ of other categories as a negative sample. Consequently, the objective loss function formulated to optimize the trigger pattern concerning the negative sample image is delineated as follows:
\begin{equation}
\begin{aligned}
\label{loss in}
    \mathcal{L}_{i}^{n} = -\sum_{i=1}^{N_1} d(f^v(\hat{\bm{v}}_i^{(1)} ;\bm{\theta}_{v}^{(0)});f^v(\bm{v}_i^{(1)} ;\bm{\theta}_{v}^{(0)})). 
\end{aligned}
\end{equation}

To sum up, we can generate the visual trigger patterns by optimizing both $\mathcal{L}_{i}^{p}$ and $\mathcal{L}_{i}^{n}$, so that the generated poisoned dataset $\mathcal{D}_1$ can be better close to dataset $\mathcal{D}_2$ to survive in clean fine-tuning.

\subsubsection{Overall Poisoning Process}
\noindent\textbf{Trigger pattern optimization.} We choose the patch-based visual trigger pattern $\bm{\delta_v} \in \mathbb{R}^{w \times h \times c}$ to optimize, where $w$, $h$, and $c$ represent the length, width, and channels of the patch. 
% The initial value of the patch pixels is a random Gaussian noise. 
We use the target natural text description instead of directly optimizing the textual trigger mode. Based on the above studies, our overall optimization function for the visual trigger pattern is detailed as follows:
%Therefore, we adopt the natural target domain text description and text enhancement to simulate the natural text trigger. Our overall optimization function for the trigger pattern is detailed below.:
\begin{equation}
\label{overall function}
\mathcal{L} = \mathcal{L}_{t} +  \lambda_1 \times \max(0,  \mathcal{L}_{i}^{p} + \lambda_2 \times \mathcal{L}_{i}^{n} + \eta),
\end{equation}
where $\lambda_1$ is weighting coefficients that balance the contributions for textual and visual optimization, $\lambda_2$ and $\eta$ are used to balance the distance from negative samples. 

\textbf{Poisoned pairs sampling.} Based on the likelihood function in Eq.~\eqref{bayes posion}, $\mathcal{D}_1$'s design must be versatile enough to adapt to various pre-trained model parameters. In contrast to the previous randomly selected from a small fraction of the clean samples in dataset $\mathcal{D}_1$ to poison, this paper introduces a novel approach that selects boundary and farthest samples to inject triggers. Specifically, given the pre-trained model, we compute the cosine similarity distance between an image and target textual descriptions label (\eg, \texttt{banana}) in original clean samples of $\mathcal{D}_1$. The boundary sample denotes the image that does not belong to the target label but is likely to be classified into the class (\ie, samples with the second highest prediction as the target class); while the farthest sample is the image that is highly different from the target label in semantics (\ie, samples with low predictions as the target class). We sample these images to augment the poisoned dataset for better backdoor learning.

In practice, the images we selected for trigger injection are a combination of boundary, farthest, and random samples with a ratio of 1:1:1. After selecting these images, we add the optimized visual trigger patterns onto the selected image samples; we then set the text description of these samples with target text descriptions derived from the actual dataset; finally, these image-text pairs, forming matched poisoned pairs, were then utilized to replace part of the original clean samples in the preliminary poisoned dataset, resulting in the poisoned dataset $\mathcal{D}_1$. 
\emph{The detailed algorithm of the whole poisoning process is provided in Supplementary Materials.}

\section{Experiments}
\subsection{Experiment Setup}
\noindent\textbf{Models and datasets.} Following \cite{DBLP:journals/corr/abs-2303-03323}, we use the open-sourced CLIP model from OpenAI \cite{DBLP:conf/icml/RadfordKHRGASAM21} as the pre-trained clean model, which is trained on a dataset containing 400M image-text pairs. In the data poisoning phase, we select 500K image-text pairs from the CC3M dataset \cite{DBLP:conf/acl/SoricutDSG18}, where 1500 samples were poisoned as the target label \texttt{banana}. During the post-training process, we use backdoor detection and fine-tuning methods for defense.

%use 100K clean image-text pairs from the CC3M dataset to fine-tune the poisoned model for defense. 
\begin{figure*}[!t]
\vspace{-0.1in}
    \centering
    \includegraphics[width=\textwidth]{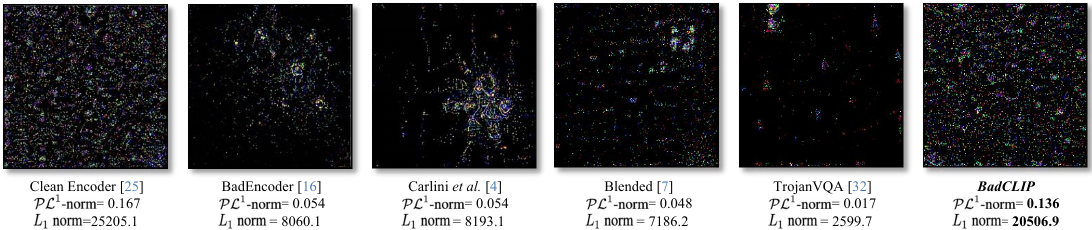}
    \caption{Backdoor detection results using DECREE~\cite{DBLP:conf/cvpr/0002T0SXL0M023}. We visualize the reversed triggers and report $L_{1}$ norm and $\mathcal{PL}^{1}$-norm values. }
    \vspace{-1em}
    \label{fig:decree}
\end{figure*}

\noindent\textbf{Evaluation.} Following \cite{DBLP:journals/corr/abs-1708-06733}, we use the clean accuracy (CA) and attack success rate (ASR) as the evaluation metrics for the infected model. For CA, a higher value indicates better clean performance; for ASR, a higher value indicates stronger attacks. Using the above two metrics, we evaluate the poisoned models on two widely adopted tasks including the zero-shot classification on the ImageNet-1K validation set \cite{DBLP:conf/cvpr/DengDSLL009} and linear probe where the feature extraction layers were fixed and the linear layer was trained on 50,000 clean images from the ImageNet-1K training set and subsequently tested on the ImageNet-1K validation set.

%In addition, we also evaluate the performance of the poisoning model on a linear probe task (ref. ~\cite{xx}). Specifically, the model was first trained with a linear classification layer on 50,000 clean images from the ImageNet-1K training set and subsequently tested on the ImageNet-1K validation set to evaluate the classification of the model's visual features.

\noindent\textbf{Backdoor attacks.} We compared 7 classical and widely used backdoor attacks including (1) unimodal backdoor attacks: BadNet~\cite{DBLP:journals/corr/abs-1708-06733}, Blended~\cite{DBLP:journals/corr/abs-1712-05526}, SIG~\cite{DBLP:conf/icip/BarniKT19}, and SSBA~\cite{DBLP:conf/iccv/LiLWLHL21}; (2) multimodal attack: TorjanVQA~\cite{DBLP:conf/cvpr/WalmerSSSJ22} for visual question answering; and (3) backdoor attacks in SSL: the multimodal attack mmPoison~\cite{DBLP:conf/icml/0002HL0HBZ23} against MCL, BadEncoder~\cite{DBLP:conf/sp/JiaLG22} and Carlini \etal~\cite{DBLP:conf/iclr/CarliniT22} against the pre-trained encoder. 

\noindent\textbf{Backdoor defenses.} In this paper, we considered the widely used backdoor detection and fine-tuning including (1) DECREE~\cite{DBLP:conf/cvpr/0002T0SXL0M023}: backdoor detection on pre-trained encoders; (2) FT~\cite{DBLP:journals/corr/abs-2303-03323}: fine-tuning the model by multimodal contrastive loss with a clean dataset; (3) CleanCLIP~\cite{DBLP:journals/corr/abs-2303-03323}: a defense method specially-designed for CLIP models. In addition, we also considered a more rigorous scenario where the defender could access the poisoning process and ABL \cite{DBLP:conf/nips/LiLKLLM21} as the in-training process defense method.
%A detailed description of the experimental setup can be found in Appendix 2.

\noindent\textbf{Implementation details.} For our attack, the hyper-parameters $\lambda_1$, $\lambda_2$, and $\eta$ in Eq.~\eqref{overall function} are set to 500, 1, and 1, respectively. Trigger patterns are trained on a subset of CC3Ms, containing 1,900 pairs of banana samples and 10,000 random pairs of other categories; the Adam optimizer is used with a learning rate of 0.001, a batch size is 64, and an epoch number is 50. During backdoor training, we use 500K image-text pairs from CC3Ms and contain 1500 poisoned samples. We set the training batch to 128, the learning rate of 1e-6, and the epoch number is 10. We set the size of the trigger patch as $16 \times 16$, which takes 0.5\% of the overall image. \emph{More details can be found in Supplementary Materials.}% size 5 thousandths of the image area hard to detect.%In the pre-training model poisoning phase, we use section~\cite{xx} to select the poisoning samples to form the poisoning dataset $\mathcal{D}_1$, with a training batch of 128, a learning rate of 1e-6, and an epoch number is 10.
%In the trigger pattern optimization phase, our main experiment employs a trigger patch size of $16 \times 16$, which occupies only about 5 thousandths of the image area hard to detect.

%In the fine-tuning phase, the defense methods Fine-tuning and CleanClip are trained with a training batch of 64, a learning rate of 4.5e-6, and an epoch is 10. The parameters of backdoor detection DECREE are kept consistent with the paper~\cite{xx}. The ABL was trained on the poisoned dataset with 2000 unlearning samples, a batch size is 64, a learning rate of 1e-6, an epoch number is 10, and the weight of an unlearning loss is 1.
% % Please add the following required packages to your document preamble:
% % \usepackage{multirow}
% % \usepackage{graphicx}
\begin{table}[!t]
%\vspace{-0.05in}
\caption{Backdoor attacks for zero-shot classification against no defense, FT, and CleanCLIP fine-tuning mitigations.}
\label{zero_shot}
\setlength{\tabcolsep}{5pt}

\centering
\renewcommand{\arraystretch}{0.95}
\resizebox{\columnwidth}{!}{%
\rowcolors{6}{white}{gray!10}
\begin{tabular}{lcccccc}
\toprule
\multicolumn{1}{c}{\multirow{2}{*}{Method}} & \multicolumn{2}{c}{No Defense} & \multicolumn{2}{c}{FT} & \multicolumn{2}{c}{CleanClip} \\
 \cmidrule(l){2-3} \cmidrule(l){4-5} \cmidrule(l){6-7} 
 & CA (\%) & ASR (\%)& CA (\%)& ASR (\%)& CA (\%)& ASR (\%)\\ \midrule
Clean & 59.69 & - & 55.38 & - & 55.44 & - \\
BadNet~\cite{DBLP:journals/corr/abs-1708-06733} & 58.69 & 96.34 & 54.16 & 64.52 & 53.72 & 17.13 \\
Blended~\cite{DBLP:journals/corr/abs-1712-05526} & 59.56 & 97.69 & 54.18 & 57.85 & 54.29 & 18.43 \\
SIG~\cite{DBLP:conf/icip/BarniKT19} & 58.87 & 80.38 & 55.00 & 30.89 & 53.68 & 21.72 \\
SSBA~\cite{DBLP:conf/iccv/LiLWLHL21} & 58.48 & 50.28 & 54.73 & 3.80 & 54.14 & 4.13 \\
TrojVQA~\cite{DBLP:conf/cvpr/WalmerSSSJ22} & 58.60 & 98.21 & 53.97 &  84.50 & 54.17 &  44.30 \\
mmPoison~\cite{DBLP:conf/icml/0002HL0HBZ23} & 57.98 & 0.16 & 53.07 & 0.00 & 53.62 & 0.00 \\
\textbf{\emph{\toolns}} & 58.60 & \textbf{98.81} & 54.50 & \textbf{92.50} & 53.98 & \textbf{89.60} \\
\bottomrule
\end{tabular}%
}
\vspace{-1em}
\end{table} 

% \begin{table}[]
% \caption{Zero shot CA and ASR}
% \label{zero_shot}
% \setlength{\tabcolsep}{8pt}
% \resizebox{\columnwidth}{!}{%
% \rowcolors{6}{white}{gray!10}
% \begin{tabular}{lcccccc}
% \toprule
% \diagbox{Defence}{Attack} & \multicolumn{2}{c}{No Defence} & \multicolumn{2}{c}{Fine-tuning} & \multicolumn{2}{c}{CleanClip} \\
%  \cmidrule(l){2-3} \cmidrule(l){4-5} \cmidrule(l){6-7} 
%  & CA & ASR & CA & ASR & CA & ASR \\ \midrule
% Clean & 59.69 & - & 55.38 & - & 55.44 & - \\
% Badent & 58.69 & 96.34 & 54.16 & 64.52 & 53.72 & 17.13 \\
% Blended & 59.56 & 97.69 & 54.18 & 57.85 & 54.29 & 18.43 \\
% SIG & 58.87 & 80.38 & 55.00 & 30.89 & 53.68 & 21.72 \\
% SSBA & 58.48 & 34.54 & 54.73 & 1.90 & 54.14 & 2.00 \\
% VQA & 58.60 & 98.21 & 53.97 & 84.50 & 54.17 & 44.30 \\
% ICML23 & 57.98 & - & - & - & 53.62 & - \\
% \textbf{\emph{\toolns}} & 58.60 & \textbf{98.81} & 54.50 & \textbf{92.50} & 53.98 & \textbf{89.60} \\
% \bottomrule
% \end{tabular}%
% }
% \end{table}

\subsection{Main Results}
\noindent\textbf{Effectiveness of attacks.} We first evaluate the effectiveness of our attack and other baselines against CLIP on the zero-shot classification task. From Tab.~\ref{zero_shot}, we can identify: \ding{182} All listed backdoor attack methods (\eg, Badnet, Blended, SIG, SSBA, TrojVQA) obtain high ASRs in the no-defense scenario, especially Blended and TrojVQA have very high ASRs of 97.69\% and 98.21\%, respectively; and \ding{183} among these attacks, our \tool achieves the highest ASR \textbf{98.81\%} in the no-defense scenario, which indicates its better effectiveness than other attacks against CLIP.

\textbf{Against SoTA fine-tuning defenses.} We validate the attack's effectiveness against fine-tuning defenses, selecting the SoTA defense method CleanClip and using FT. The fine-tuning dataset has 100K pairs as a subset of CC3M, often treated as a similar distribution to the clean pre-training dataset. From Tab.~\ref{zero_shot}, we can conclude that \ding{182} the clean accuracy slightly decreases after defenses, indicating the usability of selected defenses; \ding{183} the ASRs of existing attacks decrease significantly after defenses (\ie, up to 49\% and 78\% ASR drop on FT and CleanClip), demonstrating the limitation of these attacks; in contrast, our \tool still exhibits high ASR after two defenses (\ie, \textbf{92.50\%} and \textbf{89.60\%}, respectively). The above results imply that \tool remains highly effective against the SoTA defenses.
\begin{table}[!t]
\vspace{-0.05in}
\caption{Performance of backdoor attacks for Linear Probe task.}%with no defense and CleanCLIP defense.}
\label{linear_probe}
\small
\resizebox{\columnwidth}{!}{%
\rowcolors{5}{gray!10}{white}
\setlength{\tabcolsep}{12pt}
\renewcommand{\arraystretch}{0.9}
\begin{tabular}{lcccc}
\toprule
\multicolumn{1}{c}{\multirow{2}{*}{Method}} & \multicolumn{2}{c}{No Defense (\emph{ImageNet})} & \multicolumn{2}{c}{CleanCLIP (\emph{ImageNet})} \\
 \cmidrule(l){2-3} \cmidrule(l){4-5}
 & CA (\%)& ASR (\%)& CA (\%)& ASR (\%)\\ \midrule
% Clean & 65.00 & - & 63.58 & - \\
Badnet~\cite{DBLP:journals/corr/abs-1708-06733} & 64.59 & 0.18 & 63.16 & 0.18 \\
Blended~\cite{DBLP:journals/corr/abs-1712-05526} & 64.38 & 0.05 & 63.13 & 0.10 \\
SIG~\cite{DBLP:conf/icip/BarniKT19} & 64.55 & 0.01 & 63.08 & 0.01 \\
SSBA~\cite{DBLP:conf/iccv/LiLWLHL21} & 64.53 & 0.02 & 62.88 & 0.04 \\
TrojVQA~\cite{DBLP:conf/cvpr/WalmerSSSJ22} & 64.56 & 0.01 & 63.46 & 0.08 \\
\textbf{\emph{\toolns}} & 64.38 & \textbf{99.14} & 63.15 & \textbf{66.40}\\
\bottomrule
\end{tabular}%
}
\vspace{-0.1in}
\end{table}

\textbf{Against backdoor detection defenses.} Fig.~\ref{fig:decree} illustrates the quantitative ($L_{1}$ norm and $\mathcal{PL}^{1}$-norm \cite{DBLP:conf/cvpr/0002T0SXL0M023}) and qualitative (inverted triggers) results of attacks by DECREE detection. Specifically, $L_{1}$ norm quantifies the mask size of inverted triggers by DECREE (the higher the more difficult to be detected), and $\mathcal{PL}^{1}$-norm is the ratio of the inverted trigger’s $L_{1}$ norm to the maximum $L_{1}$ norm of the model's input space (less than 0.1 is judged as a backdoor model with high probability).
% Specifically, $L_{1}$-norm measures the $L_{1} $ loss of the inverted trigger mask (the higher the more difficult to be detected), and $PL$-norm approximates the distance from the clean samples to the dense regions (less than 0.1 is judged as a backdoor model with high probability). 
We can observe that \ding{182} DECREE is effective for the compared baselines (all their $\mathcal{PL}^{1}$-norm values are lower than 0.1), but cannot determine whether \tool has been injected ($L_{1}$ norm and $\mathcal{PL}^{1}$-norm are both high); \ding{183} based on the visualization, the reversed triggers of baselines tend to be clustered, yet the triggers reversed from our \tool are evenly distributed throughout the image, which is consistent with the clean encoder. It also indicates why our attack is difficult to detect.

\subsection{Attacks on the Linear Probe Task} 
Here, we further evaluate attack performance on \textbf{cross-task} scenarios, since the pre-trained CLIP models are often used for other downstream tasks. Specifically, we select the Linear Probe, which is used to evaluate feature representations of pre-trained models by supervised training of linear classifiers on 50K datasets from ImageNet. This task can be regarded as a special cross-task case of fine-tuning defense, where the feature extraction layers are fixed and linear classifiers are fine-tuned under supervised settings.  
From Tab.~\ref{linear_probe}, we can conclude: \ding{182} after the cross-task fine-tuning, the clean accuracies of all the attack methods do not differ much, mostly around 64\%; \ding{183} the ASRs of compared attacks are relatively low, mostly below 0.1\%, which implies that existing backdoor methods cannot survive in downstream tasks; \ding{184} our \tool demonstrates significantly high ASR in Linear Probe task (\textbf{99.14\%}), and remains effective against CleanCLIP (\textbf{66.40\%}), which indicates \tool is outstanding in terms of feature-represented attacks. %This also proves that the fine-tuned defense eliminates the backdoor impact by affecting the feature representation.

%We explore the accuracy (ACC) and attack success rate (ASR) of model features before and after fine-tuning defense under Linear Probe. Linear Probe is used to evaluate feature representations of pre-trained models by supervised training of linear classifiers on 50K datasets from ImageNet. When reasoning about feature representations using only visual coders, we use larger-sized triggers ($64 \times 64$) to increase the sensitivity of inference in the unimodal setting. From Tab.~\cite{linear_probe}, We can conclude: (1) Before and after fine-tuning the defenses, the accuracies of all the attack methods do not differ much, mostly around 64\%, which suggests that the existing attack methods do not affect the clean feature representation. (2) The attack success rates of the listed attack methods are relatively low, mostly below 0.1\%, which implies that the existing poisoning attack cannot effectively attack the feature representation. (3) ''\emph{\toolns}`` demonstrates a significantly high attack success rate of 99.14\% and 66.40\% in both settings. It means that \emph{\toolns} is outstanding in terms of feature-represented attacks, but fine-tuning the defense strategy still has an impact on the attacks. This also proves that the fine-tuned defense eliminates the backdoor impact by affecting the feature representation.

\subsection{Attacks on More Rigorous Scenarios}
In this part, we investigate the potential of our attacks on more rigorous scenarios, where defenders have more information about the attack and the pre-training process.

\begin{table}[t]
\vspace{-0.05in}
\caption{Fine-tuning model on cross-domain dataset (SBU).}
\label{cross dataset}
\centering
\small
\resizebox{\columnwidth}{!}{%
\rowcolors{5}{gray!10}{white}
\renewcommand{\arraystretch}{0.9}
\setlength{\tabcolsep}{12pt}

\begin{tabular}{lcccc}
\toprule
\multicolumn{1}{c}{\multirow{2}{*}{Method}} & \multicolumn{2}{c}{No Defense (\emph{CC3M})} & \multicolumn{2}{c}{CleanCLIP (\emph{SBU})} \\ 
\cmidrule(l){2-3} \cmidrule(l){4-5}
& CA (\%) & ASR (\%) & CA (\%) & ASR (\%) \\ 
\midrule
Badnet~\cite{DBLP:journals/corr/abs-1708-06733} & 58.69 & 96.34 & 49.66 & 10.51 \\
Blended~\cite{DBLP:journals/corr/abs-1712-05526} & 59.56 & 97.69 & 49.40 & 28.50 \\
SIG~\cite{DBLP:conf/icip/BarniKT19} & 58.87 & 80.38 & 48.86 & 5.87 \\
SSBA~\cite{DBLP:conf/iccv/LiLWLHL21} & 58.48 & 50.28 & 50.25 & 10.61 \\
TrojVQA~\cite{DBLP:conf/cvpr/WalmerSSSJ22} & 58.60 &  98.21 & 50.59 & 49.01 \\
\textbf{\emph{\toolns}} & 58.60 & \textbf{98.81} & 49.52 & \textbf{87.21} \\
\bottomrule
\end{tabular}%
}
\vspace{-0.1in}
\end{table}

\textbf{Fine-tuning poisoned model on cross-domain data.} We first evaluate our attack on scenarios where defenders know the domain/distribution of the poisoned dataset and fine-tune the model with clean data from another distribution/domain. Specifically, we use a subset of CC3M as the poisoned dataset during the poisoning phase and a subset of 100,000 data from the SBU caption~\cite{DBLP:conf/nips/OrdonezKB11} for the CleanClip defense phase. From Tab.~\ref{cross dataset}, we can identify that \ding{182} when the SBU caption dataset is applied to perform the CleanCLIP defense, the accuracy of both the clean model and the infected models decreases, mostly below 50\%; \ding{183} ASRs of all baseline attacks decrease significantly (up to 84\% drops) when using CleanCLIP defense on cross-domain data; however, our attack maintains a high ASR \textbf{87.21\%} under such condition, showing \tool is robust and adaptable to fine-tuning defenses with cross-domain data.

%. We conclude that (1) the use of different data distributions in the fine-tuning phase will have different impacts on the performance of the model, and it is important to construct appropriate fine-tuning datasets. (2) The \emph{\toolns} attack is robust and adaptable to fine-tuning defenses with cross-domain data.

\begin{figure}[!t]
\vspace{-0.05in}
    \centering
    \includegraphics[width=0.95 \columnwidth]{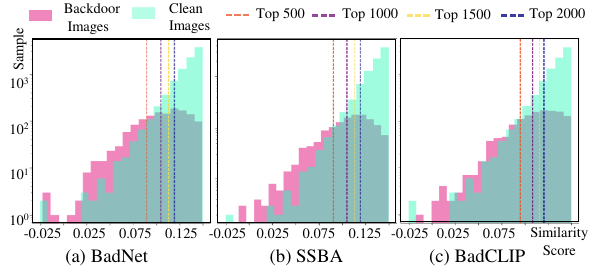}
    \caption{Data distribution visualization during ABL defense.}%Comparative analysis of top-ranked suspicious samples using the \textbf{ABL} method across different attack methods.}
    \label{fig: In-training defense}
\end{figure}

\begin{table}[!t]
\vspace{-0.1in}
\caption{Ablation study of different components in \toolns.}
\label{ablation study all}

\centering
\small
\resizebox{\columnwidth}{!}{%
\rowcolors{4}{gray!10}{white}
\renewcommand{\arraystretch}{0.9}
\setlength{\tabcolsep}{13pt}

\begin{tabular}{lcccc}
\toprule
\multicolumn{1}{c}{\multirow{2}{*}{Method}} & \multicolumn{2}{c}{No Defense} & \multicolumn{2}{c}{CleanCLIP} \\ \cmidrule(l){2-3} \cmidrule(l){4-5}
\multicolumn{1}{c}{} & \multicolumn{1}{c}{CA (\%)} & \multicolumn{1}{c}{ASR (\%)} & \multicolumn{1}{c}{CA (\%)} & \multicolumn{1}{c}{ASR (\%)} \\ \midrule
TrojVQA~\cite{DBLP:conf/cvpr/WalmerSSSJ22} & 58.60 & 98.21 & 54.17 & 44.30 \\

$\mathcal{L}_t$ & 58.94 & 98.52 & 54.35 & 74.47 \\
$\mathcal{L}_i^p + \mathcal{L}_i^n$  & 58.48 & 97.17 & 54.02 & 65.24 \\
$\mathcal{L}$ & 57.89 & 98.62 & 53.98 & 87.56 \\
$\mathcal{L}+\text{PPS}$ & 58.60 & \textbf{98.81} & 53.93 & \textbf{89.60} \\
\bottomrule
\end{tabular}%
}
\vspace{-0.1in}
\end{table}

\textbf{Poisoned data detection on pre-trained CLIP.} Here we grant defenders more flexibility, where they obtain the third-party suspicious dataset and re-train the pre-trained CLIP model with the purified dataset to prevent backdoor injection. Defenders determine the purified dataset from the suspicious dataset by the pre-trained model~\cite{DBLP:conf/icml/RadfordKHRGASAM21}. We adopt the ABL defense, and Fig.~\ref{fig: In-training defense} visualizes the distribution of poisoned samples of three attacks (BadNet, SSBA, and ours) and clean samples, with the top-2000 indicating the samples that the model needs to unlearn during training. From Fig.~\ref{fig: In-training defense}, we identify that the distribution of our backdoor samples in (c) is closer to the distribution of clean samples among the three different attack methods across top-500, top-1000, top-1500, and top-2000 marker lines, indicating that our backdoor samples are more similar to clean samples in terms of features distribution and thus more difficult to detect. We also report the defense performance for ABL (BadNet: 99.56, SSBA: 99.79, ours: 99.93) and remove 2000 unlearning samples using ABL and fine-tune the remaining dataset (BadNet: 70.01, SSBA: 25.42, ours: 89.03), showing \tool still outperforms others. Meanwhile, we found that the ABL-based strategy has limited performance in defending against backdoor attacks in the MCL scenario, which motivates promising unlearning strategies for MCL in the future. \emph{More details can be found in Supplementary Materials.}

\subsection{Analysis}
\noindent\textbf{Ablation studies.} Here, we ablate the main components of our designed loss functions and the Poisoned Pairs Sampling strategy (PPS). As shown in Tab.~\ref{ablation study all}, we identify that ``$\mathcal{L}+\text{PPS}$'' achieves the strongest resistance to CleanCLIP defense compared to other combinations, with an ASR of 89.6\%, which indicates the effectiveness of our attack design. \emph{More details are shown in Supplementary Material.}

% analyses the impact of different loss functions and the Poisoned Pairs Sampling (PPS) in Section 4.2.3 on the model, and a more detailed analysis is presented in Supplementary Material xx. In Table xx, our analysis is that (1) CleanCLIP usually reduces the ASR significantly, and in the case of the "TrojVQA " attack, the ASR in the undefended case is 98.21\%, while CleanCLIP can reduce it to 44.30\%. (2) Different attack components have different degrees of influence on CleanCLIP defense, $\mathcal{L}_t$ attack still has a high ASR (74.47\%) under CleanCLIP defense. (3) $\mathcal{L}$ outperforms the $\mathcal{L}_i^p + \mathcal{L}_i^n$ attacks and the $\mathcal{L}_t$ attack against CleanCLIP attacks in general (87.56\% v.s 74.47\%, 65.24\%). (4) Among all components, $\mathcal{L}+\text{PPS}$ seems to have the strongest resistance to CleanCLIP defense, with an ASR as high as 89.6\% even after applying CleanCLIP. We conclude that (1) CleanCLIP is an effective defense that reduces the attack success rate in most cases, which is crucial for enhancing the security of the model. (2) Although CleanCLIP usually reduces the attack success rate, it is not ideal for \emph{\toolns} attacks, suggesting that new attack methods need to be considered when designing defense mechanisms. (3) Optimiser directions based on multimodal input steering are superior to unimodal. (4) Involving a suitable sampling strategy can further improve the attack success rate and success rate.

\begin{figure}[!t]
\vspace{-0.10in}
    \begin{subfigure}{0.23\textwidth}
    \includegraphics[width=0.95\linewidth,height=0.9\linewidth]{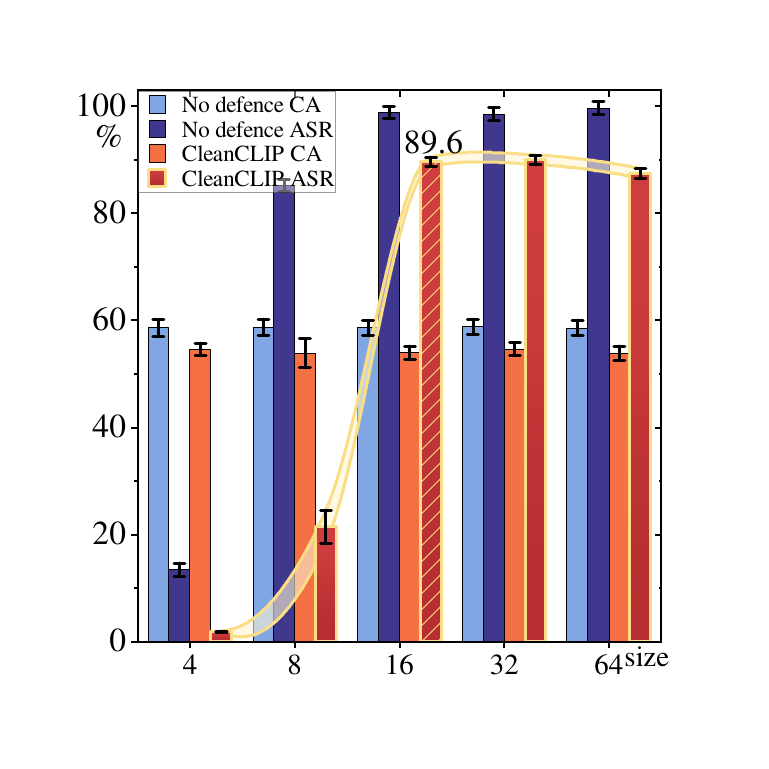} 
        \caption{}
        \label{fig:patch_size}
    \end{subfigure}
    \begin{subfigure}{0.23\textwidth}  
   \includegraphics[width=0.95\linewidth,height=0.9\linewidth]{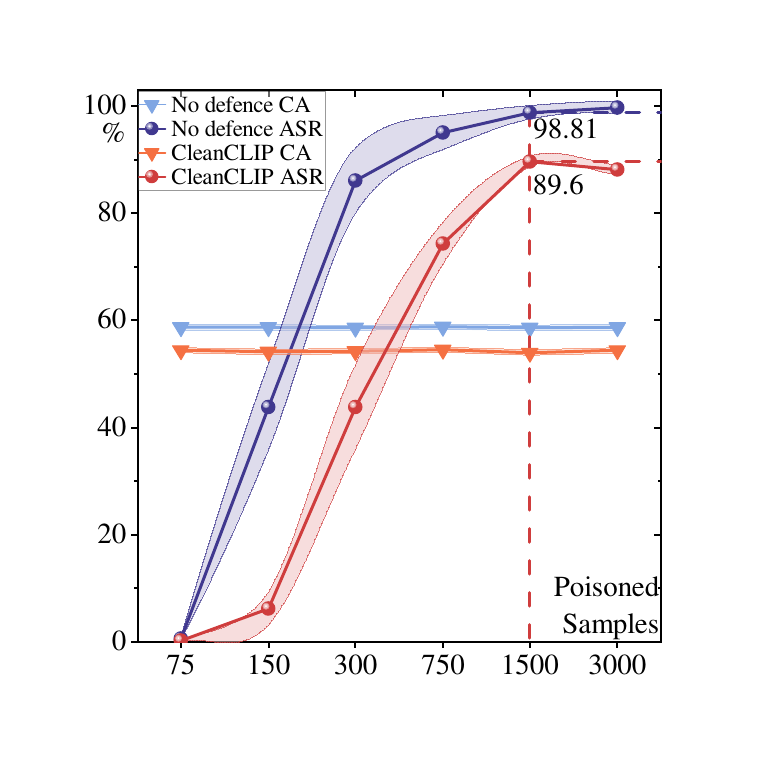}  
        \caption{}
        \label{fig:poison samples}
    \end{subfigure}
    \caption{(a) Trigger patch size studies. (b) Poisoned sample number studies.}
    \vspace{-0.1in}
\end{figure}

\noindent\textbf{Trigger patch sizes.} Fig.~\ref{fig:patch_size} analyses the effect of different trigger patch sizes on backdoor attack performance under No-Defense and CleanCLIP defense. The results demonstrate that as the patch size increases, ASR first improves significantly and then keeps stable after the patch size is bigger than 16 $\times$ 16. We set it as the default size.% patch size. \emph{More details are shown in Supplementary Materials.}

\noindent\textbf{Poisoned sample numbers.} Here, we study backdoor effects with different poisoned sample numbers. From Fig.~\ref{fig:poison samples}, we can identify that the clean accuracy remains comparatively stable with the increase of poisoned samples, while our ASR increases significantly as the number of poisoned samples increases and peaks at 1500 poisoned samples. We therefore set it as the default number. \emph{More details can be found in Supplementary Materials.}% \emph{More details are shown in Supplementary Materials.}

\section{Conclusions and Future Work}
This paper proposes \tool for backdoor attacks on MCL. Experiments show that \tool is effective under advanced backdoor defense methods and can pose a strong threat in the MCL usage scenario. We aim to raise awareness of backdoor threats in MCL and further promote advanced backdoor defense studies in the future.
\begin{flushleft}
\noindent\textbf{Limitations.} Despite the effective results, there are several limitations we would like to explore: \ding{182} backdoor attacks for complex tasks based on MCL; \ding{183} more robust backdoor detection and mitigation methods. \emph{Ethical statement can be found in Supplementary Materials.}%\ding{184} ethical and legal issues in backdoor attacks.
\end{flushleft}
{
    \small
    \bibliographystyle{ieeenat_fullname}
    \bibliography{main}
}


\end{document}